\documentclass[a4paper,twoside]{article}

\usepackage{epsfig}
\usepackage{subcaption}
\usepackage{calc}
\usepackage{amssymb}
\usepackage{amstext}
\usepackage{amsmath}
\usepackage{amsthm}
\usepackage{multicol}
\usepackage{pslatex}
\usepackage{apalike}
\usepackage{algorithm2e}
\usepackage{hyperref}
\usepackage[bottom]{footmisc} 
\usepackage{makecell}
\usepackage{SCITEPRESS}
\usepackage{multirow}
\begin{document}

\title{Predicting Stock Price Movement with LLM-Enhanced Tweet Emotion Analysis}

\author{\authorname{An Vuong\sup{1} and Susan Gauch\sup{1} }
\affiliation{\sup{1}Department of Electrical Engineering and Computer Science, University of Arkansas, 1 University of Arkansas, Fayettville, Arkansas, United States of America}
\email{\{anv, sgauch\}@uark.edu}
}

\abstract{
Accurately predicting short-term stock price movement remains a challenging task due to the market’s inherent volatility and sensitivity to investor sentiment. This paper discusses a deep learning framework that integrates emotion features extracted from tweet data with historical stock price information to forecast significant price changes on the following day. We utilize Meta’s Llama 3.1-8B-Instruct model to preprocess tweet data, thereby enhancing the quality of emotion features derived from three emotion analysis approaches: a transformer-based DistilRoBERTa classifier from the Hugging Face library and two lexicon-based methods using National Research Council Canada (NRC) resources. These features are combined with previous-day stock price data to train a Long Short-Term Memory (LSTM) model. Experimental results on TSLA, AAPL, and AMZN stocks show that all three emotion analysis methods improve the average accuracy for predicting significant price movements, compared to the baseline model using only historical stock prices, which yields an accuracy of 13.5\%. The DistilRoBERTa-based stock prediction model achives the best performance, with accuracy rising from 23.6\% to 38.5\% when using LLaMA-enhanced emotion analysis. These results demonstrate that using large language models to preprocess tweet content enhances the effectiveness of emotion analysis which in turn improves the accuracy of predicting significant stock price movements.
}

\onecolumn \maketitle \normalsize \setcounter{footnote}{0} \vfill

\section{\uppercase{Introduction}}
\label{sec:introduction}
\noindent
Financial time series forecasting, particularly stock price prediction, has long been one of the most challenging problems for researchers and investors. With the U.S. Securities and Exchange Commission officially transitioning the trade settlement cycle from T+2 to T+1 on May 28, 2024~\cite{sec2024-62}, short-term trading strategies that involve opening and closing positions within the same day or by the following day, have become increasingly common as traders seek to capture quick profits from significant price movements. This shift, along with the market’s inherent complexity and sensitivity to factors such as corporate news, macroeconomic indicators, and investor sentiment, underscores the growing importance of accurate stock price movement prediction to support timely and informed trading decisions.

In recent years, researchers have increasingly investigated social media as a source of predictive signals for financial markets. For instance, Bollen et al.~\cite{Bollen2011} and Nguyen et al.~\cite{Nguyen2015} demonstrated that aggregate sentiment from platforms such as Twitter and investor forums can enhance short-term stock movement prediction. While traditional sentiment analysis typically categorizes text into \textit{positive}, \textit{negative}, or \textit{neutral} \cite{Wankhade2022}, recent studies have shifted toward emotion detection from text to capture more nuanced linguistic signals across diverse domains, including management and marketing, healthcare, education, public monitoring ~\cite{Kusal2022}. In financial contexts, emotions such as \textit{fear}, \textit{joy}, \textit{sadness}, and \textit{stress} extracted from social media textual data have been shown to significantly influence short-term returns and volatility in major indices like the S\&P~500~\cite{Griffith2020}. Although prior studies such as \cite{Chun2022} have explored the use of emotion features in financial forecasting, the integration of such features into stock price movement prediction models remains relatively underexplored.

Our research developed a deep learning framework that leverages emotion analysis of social media content to predict significant next-day stock price movements. In this approach, features extracted from the previous day are used to train the model in a supervised learning setting to classify whether the stock will significantly increase, decrease, or remain stable on the next day.

The main contributions of this study are as follows: 
(1) we leverage a large language model (Meta's Llama 3.1) to preprocess tweet content before applying emotion analysis, which improves the quality of extracted emotional features; 
(2) we propose a novel framework for stock price significant movement prediction that integrates emotion features derived from tweet data with daily stock price data; and 
(3) We systematically compare the predictive performance of a baseline model that uses only historical stock prices with three models that combine stock price data and emotion features from 3 different emotion analysis methods: the DistilRoBERTa-based Emotion Classification Model, the NRC Emotion Intensity Lexicon, and the NRC Emotion Label Lexicon. Each emotion analysis method is evaluated with and without LLaMA-enhanced emotion analysis.

The remainder of this paper is organized as follows. Section~2 reviews related work on stock prediction and the use of sentiment and emotion analysis in financial contexts. Section 3 describes our proposed methodology, including LLaMA-based tweet preprocessing, emotion analysis on tweet data, and the construction of the final dataset to predict significant stock price movements. Section 4 presents the experimental results and evaluates the effectiveness of three emotion analysis methods, both with and without LLaMA-enhanced emotion analysis, in improving the accuracy of predicting significant stock price movements. Finally, Section 5 concludes the paper and discusses possible directions for future work.

\section{\uppercase{Related Work}}
\noindent
Stock market prediction remains a highly active area of research in both academia and industry. Over the decades, numerous predictive models have been proposed to address the inherent complexity and non-linearity of financial time series.

In the 1990s, artificial neural networks (ANNs) were the most widely used models for stock market prediction~\cite{Atsalakis2009}. For instance, Kimoto et al.~\cite{Kimoto1990} implemented a modular neural network model using historical stock prices, technical indicators, and macroeconomic variables as input features to predict one-month-ahead movements of the Tokyo Stock Price Index (TOPIX). In the following decade, support vector machines (SVM) ~\cite{Huang2005} and support vector regression (SVR)~\cite{Huang2009} emerged as popular alternatives, offering improved generalization and robustness by leveraging the structural risk minimization principle. Specifically, Huang et al.~\cite{Huang2005} applied a support vector machine (SVM) to predict the directional movement of the NIKKEI 225 Index using macroeconomic data, and demonstrated that SVM outperformed artificial neural networks (ANNs) in classification accuracy. In the past decade, deep learning techniques have gained increasing attention in financial forecasting. Among them, Long Short-Term Memory (LSTM) networks and their variants have been widely adopted for stock market prediction ~\cite{Jiang2021}. A particularly representative work is that of Nelson et al.~\cite{Nelson2017}, who applied an LSTM-based model to predict stock price movement of Brazilian stocks and demonstrated that it achieved significantly higher accuracy compared to four traditional machine learning models.

Recent studies have increasingly incorporated external textual sources such as financial news, social media, and web searches. A common method to process this unstructured data is sentiment analysis, which determines whether the content reflects a positive or negative outlook on the market ~\cite{Balaji2017}. In 2011, Bollen et al.~\cite{Bollen2011} used sentiment analysis to extract mood from Twitter data and integrated these features with daily DJIA-closing values to predict the movement of the Dow Jones Industrial Average (DJIA). Similarly, Nguyen et al.~\cite{Nguyen2015} proposed a hybrid model that integrates sentiment features extracted from financial message boards with lagged stock prices to predict daily stock movement.

In recent years, emotion analysis has been widely applied to domains such as healthcare, education, marketing, and finance, with a primary focus on analyzing text from online social media, review systems, and conversational agents~\cite{Kusal2022} ~\cite{Kusal2022}. For example, Mackey et al.~\cite{mackey2021fakenews} proposed a fake news detection framework that combines emotional features extracted from news articles with BERT embeddings. Their model incorporates discrete emotion vectors such as anger, trust, and joy, as well as continuous emotion dimensions including valence and arousal, to improve the classification of misinformation into categories such as satire, hoax, propaganda, and clickbait. In the context of stock market prediction, the application of text-based emotion analysis remains relatively limited. Among the few existing studies, Chun et al.~\cite{Chun2022} proposed an emotion-based stock prediction system that integrates multiple emotional categories including joy, interest, surprise, fear, anger, sadness, and disgust, extracted from investor microblogs. Their model was designed to predict the daily directional movement of the KOSPI 200 index futures based on these multidimensional emotional indicators.

\section{\uppercase{Approach}}
\begin{figure*}[htbp]
    \centering
    \includegraphics[width=\textwidth]{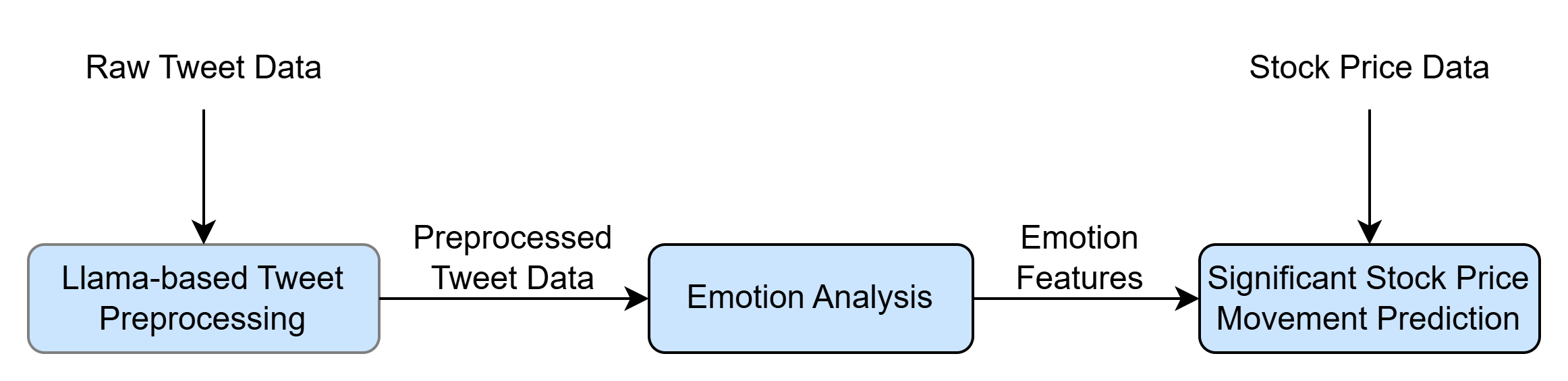}
    \caption{The overall pipeline for predicting significant stock price movements}
    \label{fig:pipeline}
\end{figure*}

\noindent 
In this section, we present our framework for predicting significant stock price movements by integrating emotion analysis results from tweet content with historical stock price data. Figure~\ref{fig:pipeline} illustrates that our approach comprises three main components: (1) preprocessing stock-related tweets using a large language model, (2) generating three sets of emotion features using a transformer-based classifier and two lexicon-based methods, and (3) integrating the extracted emotion features with historical stock prices to construct the final dataset, which is then used to train an LSTM model to predict the next day's significant stock price movement.

\subsection{Llama-based Tweet Preprocessing}
\noindent
Let $\mathcal{D} = \{x_1, x_2, ..., x_N\}$ denote a dataset of $N$ tweets, where each $x_i$ is a raw tweet content. To extract emotional information from these tweets, we employ a prompt-based querying mechanism using the Meta’s Llama 3.1-8B-Instruct model~\cite{Llama3.1}. For each tweet $x_i \in \mathcal{D}$, we construct a prompt $x_i^{\text{prompt}}$ that is passed into the LLM to produce a set of predicted emotion annotations, denoted by $\hat{y}_i^{\text{LLM}}$:

\[
\hat{y_i}^{\text{LLM}} = \text{LLM}({x_i}^{\text{prompt}})
\]

\noindent
\textbf{Prompt Template} Each tweet \( x_i \) is passed into the following prompt template to create the final input \( x_i^{\text{prompt}} \) for the LLaMA model:

\begin{quote}
\textit{You will be given a human-written tweet. Identify all possible emotions expressed in the tweet.  Return the output as a comma-separated list of emotion-related words that are relevant to the stock market context. If no emotion is detected, return "no emotion".}
\end{quote}

\noindent
Tweets that return “no emotion” are removed (LLaMA-based emotion filtering). The remaining tweets are then further preprocessed using the NLTK toolkit, including converting text to lowercase, removing stop words and punctuation. Examples of LLaMA-based tweet preprocessing are provided in Table~\ref{tab:Llama_emotion_examples}.

\begin{table*}[t]
\centering
\caption{Examples of Llama-based tweet preprocessing}
\label{tab:Llama_emotion_examples}
\begin{tabular}{|p{10cm}|p{3.5cm}|}
\hline
\textbf{Tweet} & \textbf{Extracted Emotions} \\
\hline
\textit{CPI numbers drop tomorrow. If it comes in soft, TSLA is gonna explode. Loaded up today.} & anticipation, excitement, confidence \\
\hline
\textit{Feeling uneasy about tomorrow’s Fed meeting. Already trimmed some AAPL just in case.} & anxiety, fear, caution \\
\hline
\textit{MSFT Q2 report is scheduled for Thursday after market close.} & no emotion \\
\hline
\end{tabular}
\end{table*}

\subsection{Emotion Analysis}
\noindent
In this study, we explore three emotion analysis methods applied to stock-related tweets. Each method produces a different emotion representation, which is then combined with daily stock price data to form multiple versions of the final dataset.

\textbf{Method 1} (DistilRoBERTa) utilizes a pre-trained DistilRoBERTa-based model available through the Hugging Face library~\cite{hartmann2022emotion}. For each tweet, the model returns probability scores across seven emotions. These scores represent the model’s confidence in the presence of each emotional state and form the following seven-dimensional emotion vector:

\begin{align*}
E_1 = \{ \mathit{anger},\ \mathit{disgust},\ \mathit{neutral},\ \mathit{fear},\ \mathit{joy},\mathit{sadness},\\
\phantom{E_1 = \{}  \mathit{surprise} \}
\end{align*}

\textbf{Method 2} (NRC-Intensity) and \textbf{Method 3} (NRC-Binary) are lexicon-based methods that generate tweet-level emotion vectors using the National Research Council Canada (NRC) emotion resources~\cite{mohammad2013}. We follow the emotion vector construction approach described in Mackey et al.~\cite{mackey2021fakenews}, where each text instance is represented by aggregating lexicon-based emotion scores of matched tokens. In this work, each tweet is tokenized and matched against the corresponding lexicon to compute an emotion vector.

In \textbf{Method 2}, each token is assigned an intensity score ranging from 0 to 1 across the following eight emotions:

\begin{align*}
E_2 = \{ \mathit{anger},\ \mathit{anticipation},\ \mathit{disgust},\ \mathit{fear},\ \mathit{joy},\\
\phantom{E_1 = \{} \mathit{sadness}, \ \mathit{surprise},\ \mathit{trust} \}
\end{align*}

In \textbf{Method 3}, each token is assigned a binary score (0 or 1) for ten emotion categories, including two additional polarity dimensions:

\begin{align*}
E_3 = \{ \mathit{anger},\ \mathit{anticipation},\ \mathit{disgust},\ \mathit{fear},\ \mathit{joy},\\
\phantom{E_3 = \{} \mathit{sadness},\ \mathit{surprise},\ \mathit{trust},\ \mathit{positive},\ \mathit{negative}\}
\end{align*}

\noindent
For both methods, the normalized score for each emotion \(i\) is calculated as:

\[
\hat{e}_i = \frac{1}{|M|} \sum_{w \in T} s(w, i),
\]

\noindent
where \(T\) is the set of tokens in the tweet, \(M\) is the subset of tokens matched to at least one emotion, and \(s(w, i)\) represents either the intensity score (Method 2) or the binary score (Method 3) for word \(w\) and emotion \(i\).

\begin{figure}
    \centering
    \includegraphics[width=1\linewidth]{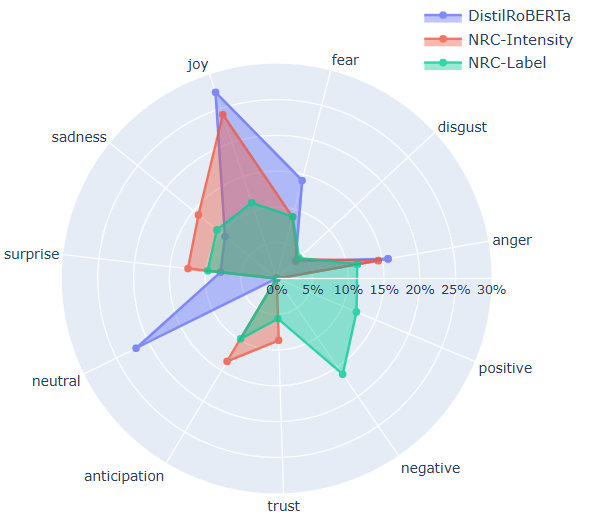}
    \caption{Distribution of extracted emotions from tweets using three different emotion analysis methods}
    \label{fig:enter-label}
\end{figure}

\subsection{Significant Stock Price Movement Prediction}

After computing tweet-level emotion scores, we aggregate them by day to obtain the average probability of each emotion. These daily averages reflect how strongly and frequently each emotion is expressed across tweets, providing insight into the dominant emotional tone of the day. This information serves as a valuable signal for predicting stock price movements. Additionally, we record the number of raw tweets before Llama-based filtering to capture daily public attention. A higher tweet volume may indicate increased market interest and a greater likelihood of significant price movement.

\noindent 
Historical stock price data is collected from Yahoo Finance~\cite{yahoo_finance} to align with the tweet dataset.
To capture daily price volatility of each stock, we calculate the daily percentage change of the closing price as follows:

\[
PC_t = \frac{P_t - P_{t-1}}{P_{t-1}} \times 100
\]

\noindent
where \(PC_t\) is the percentage change of the closing price at day \(t\), \(P_t\) is the closing price at day \(t\), and \(P_{t-1}\) is the closing price on the previous day \(t-1\).

Each stock price movement is then classified into one of three classes based on the standard deviation (\(\sigma\)) of its percentage changes:

\begin{itemize}
    \item \textbf{Stable}: if the daily percentage change is within \([- \sigma, + \sigma]\)
    \item \textbf{Significant Increase}: if the daily percentage change exceeds \(+ \sigma\)
    \item \textbf{Significant Decrease}: if the daily percentage change is lower than \(- \sigma\)
\end{itemize}

\begin{figure*}[t]
    \centering
    \includegraphics[width=0.95\textwidth]{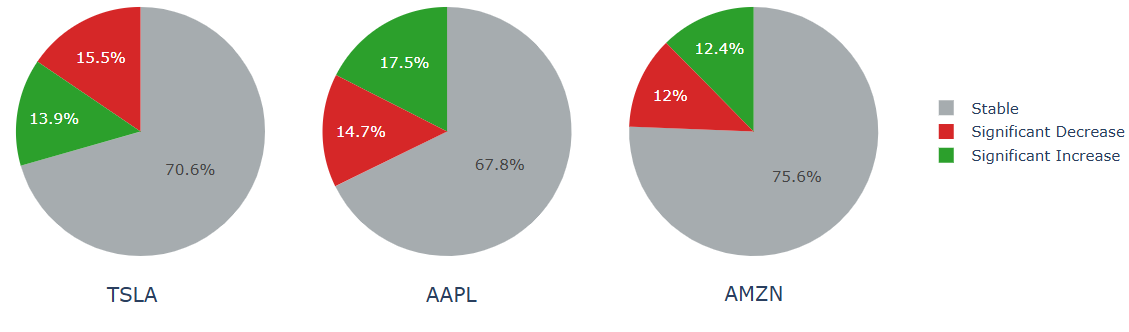}
    \caption{Class distribution of daily significant stock price movements for TSLA, AAPL, and AMZN}
    \label{fig:stock_change_pie}
\end{figure*}

After labeling, each day’s stock price data is merged with the corresponding emotion features, including the average emotion scores and tweet volume. The final dataset consists of the following features:

\begin{itemize}
    \item \textbf{Date}
    \item \textbf{Change Level}: stock price movement label (Stable, Significant Increase, Significant Decrease)
    \item \textbf{Stock Price Features}: open price, close price, high price, low price, volume
    \item \textbf{Emotion Features}: daily average emotion scores and tweet volume
\end{itemize}

We formulate the prediction of significant stock price movements as a multi-class classification task, where the goal is to categorize daily stock movement into three classes: \textit{Stable}, \textit{Significant Increase}, and \textit{Significant Decrease}.
\noindent 
Following prior studies, we adopt Long Short-Term Memory (LSTM) networks due to their effectiveness in modeling sequential dependencies and non-linear patterns in financial time series data \cite{Jiang2021}.

\noindent 
The LSTM model consists of two stacked LSTM layers with 128 hidden units each, followed by Dropout layers with a rate of 0.2 to prevent overfitting. The final Dense layer with softmax activation outputs the probability distribution across the three movement classes: Stable, Significant Increase, and Significant Decrease. The model is trained using the Adam optimizer with a learning rate of 0.01 for 200 epochs and a batch size of 32.

\section{\uppercase{Experiments}}

\subsection{Dataset}

In this study, we used stock-related tweets collected from an open-source dataset on Kaggle~\cite{equinxx2022stocktweets}, covering the period from September 30, 2021 to September 30, 2022. The dataset includes tweets associated with three prominent U.S. technology stocks: Tesla (TSLA), Apple (AAPL), and Amazon (AMZN). Each tweet entry contains the tweet content, posting date, stock ticker, and company name. Table~\ref{tab:tweet_volume_filtering} presents the number of tweets of each stock before and after applying Llama-based emotion filtering.

\begin{table}[htbp]
\centering
\caption{Tweet Volume Before and After Llama-Based Filtering}
\label{tab:tweet_volume_filtering}
\begin{tabular}{|c|c|c|}
\hline
\makecell{\textbf{Stock} \\ \textbf{Name}} & \makecell{\textbf{Tweets Volume} \\ \textbf{Before Filtering}} & \makecell{\textbf{Tweets Volume} \\ \textbf{After Filtering}} \\
\hline
TSLA  & 37173 & 18080 \\
\hline
AAPL  & 5033  & 2046  \\
\hline
AMZN  & 4077  & 1659  \\
\hline
\end{tabular}
\end{table}

The second data source is historical daily stock price data for TSLA, AAPL, and AMZN, collected from Yahoo Finance. This data includes five columns: open, close, high, low prices, and trading volume. The stock price data corresponds to the same period as the tweet dataset. After applying emotion analysis to the emotion labels extracted by the Llama-based tweet preprocessing, the results were merged with corresponding stock price data, we constructed a combined dataset covering the period from September 30, 2021 to September 30, 2022.

\subsection{Experiment Setup}

We conduct experiments under two main settings to evaluate the contribution of emotion features in predicting significant stock price movements. First, we establish a baseline by training an LSTM model using only historical stock price features. Next, we augment the model with emotion features derived from three different emotion analysis methods: (1) a DistilRoBERTa-based Emotion Classification Model (DistilRoBERTa), (2) NRC Emotion Intensity Lexicon (NRC-Intensity), and (3) NRC Emotion Label Lexicon (NRC-Lexicon). Each method is evaluated under two different stock prediction methods:

\begin{itemize}
    \item \textbf{Emotion Analysis}: Emotion features are extracted from tweets that have been preprocessed using standard text cleaning techniques (e.g., tokenization, stopword removal with NLTK toolkit).
    \item \textbf{Llama-enhanced Emotion Analysis}: Tweets are first preprocessed using Meta’s Llama 3.1-8B-Instruct model to extract emotion labels and then remove tweets with no emotional content. The remaining tweets are then further cleaned using the same NLTK preprocessing before applying emotion analysis.
\end{itemize}

In both stock prediction methods, emotion features, tweet volume before emotion-based filtering, and stock price features from the previous trading day are used to predict the movement class (Stable, Significant Increase, or Significant Decrease) for the following day. This design choice is motivated by prior findings that the predictive power of tweet-based sentiment signals decays rapidly over time, with one-day lagged sentiment exhibiting the strongest correlation with stock price movements~\cite{Teti2019}.

The combined dataset covers the period from September 30, 2021 to September 30, 2022, which includes 250 trading days. We split the data chronologically into 70\% for training (175 days) and 30\% for testing (75 days). Each experimental configuration is run 10 times to account for the randomness in LSTM training, and we report the average results to ensure consistency and robustness.

To evaluate the classification performance of our models, we use four metrics, each capturing the model’s ability to detect a specific type of stock price movement:

\begin{itemize}
    \item \textbf{Significant Increase Accuracy (S-I)}: Accuracy in predicting days with a significant upward price movement.
    \item \textbf{Significant Decrease Accuracy (S-D)}: Accuracy in predicting days with a significant downward price movement.
    \item \textbf{Stable Accuracy}: Accuracy in identifying days when the stock price remains within a stable range.
    \item \textbf{Average S-I and S-D Accuracy}: The average accuracy of predicting significant increase and significant decrease movements, reflecting the model’s overall ability to capture high-volatility movements.
\end{itemize}

\subsection{Experimental Results}

Table~\ref{tab:emotion_stock_price_results} summarizes the result of the stock prediction method using emotion analysis methods , including DistilRoBERTa, NRC-Intensity, and NRC-Label. The results show that incorporating emotion features improves the model's ability to detect high-volatility movements. Specifically, the overall average of S-I and S-D accuracy across all three stocks increases from 13.5\% (baseline) to 23.6\% (DistilRoBERTa), 18.8\% (NRC-Intensity), and 23.1\% (NRC-Label). While this improvement results in a decrease in stable accuracy, from 81.9\% (baseline) to 57.4\%, 58.2\%, and 57.8\% respectively, all models using emotion features still demonstrate strong overall performance. Among them, DistilRoBERTa achieves the best trade-off between detecting volatile movements and maintaining stability.

\begin{table}[htbp]
\centering
\scriptsize
\caption{Stock Price Prediction Results With Emotion Analysis}
\setlength{\tabcolsep}{3.5pt}
\renewcommand{\arraystretch}{1.0}
\begin{tabular}{|c|l|c|c|c|c|}
\hline
\makecell{\textbf{Stock} \\ \textbf{Name}} & \makecell[l]{\textbf{Emotion}  \\ \textbf{Analysis} \\ \textbf{Method}} & \textbf{S-I} & \textbf{S-D} & \textbf{Stable} & \makecell[c]{\textbf{Average} \\ \textbf{S-I} \\ \textbf{\& S-D}}  \\
\hline
\multirow{4}{*}{TSLA}
& Baseline & 17.5\% & 1.7\% & 82.5\% & 9.6\% \\
& DistilRoBERTa & 16.2\% & 8.3\% & 62.5\% & 12.3\% \\
& \textbf{NRC-Intensity} & \textbf{26.2\%} & \textbf{8.3\%} & \textbf{71.8\%} & \textbf{17.3\%} \\
& NRC-Label & 23.8\% & 6.7\% & 74.3\% & 15.2\% \\
\hline
\multirow{4}{*}{AAPL}
& Baseline & 10.7\% & 0.0\% & 97.0\% & 5.4\% \\
& DistilRoBERTa & 16.7\% & 28.0\% & 60.2\% & 22.3\% \\
& NRC-Intensity & 18.6\% & 0.0\% & 66.0\% & 9.3\% \\
& \textbf{NRC-Label} & \textbf{36.4\%} & \textbf{27.5\%} & \textbf{59.2\%} & \textbf{32.0\%} \\
\hline
\multirow{4}{*}{AMZN}
& Baseline & 50.8\% & 0.0\% & 66.1\% & 25.4\% \\
& \textbf{DistilRoBERTa} & \textbf{32.5\%} & \textbf{40.0\%} & \textbf{49.4\%} & \textbf{36.3\%} \\
& NRC-Intensity & 35.0\% & 24.4\% & 36.9\% & 29.7\% \\
& NRC-Label & 30.0\% & 14.0\% & 39.8\% & 22.0\% \\
\hline
\multirow{4}{*}{Average}
& Baseline & 26.3\% & 0.6\% & 81.9\% & 13.5\% \\
& \textbf{DistilRoBERTa} & \textbf{21.8\%} & \textbf{25.4\%} & \textbf{57.4\%} & \textbf{23.6\%} \\
& NRC-Intensity & 26.6\% & 10.9\% & 58.2\% & 18.8\% \\
& NRC-Label & 30.1\% & 16.1\% & 57.8\% & 23.1\% \\
\hline
\end{tabular}
\label{tab:emotion_stock_price_results}
\end{table}

\begin{table}[htbp]
\centering
\scriptsize
\caption{Stock Price Prediction Results With Llama-Enhanced Emotion Analysis}
\setlength{\tabcolsep}{3.5pt}
\renewcommand{\arraystretch}{1.0}
\begin{tabular}{|c|l|c|c|c|c|}
\hline
\makecell{\textbf{Stock} \\ \textbf{Name}} & \makecell[l]{\textbf{Emotion}  \\ \textbf{Analysis} \\ \textbf{Method}} & \textbf{S-I} & \textbf{S-D} & \textbf{Stable} & \makecell[c]{\textbf{Average} \\ \textbf{S-I} \\ \textbf{\& S-D}}  \\
\hline
\multirow{4}{*}{TSLA}
& Baseline & 17.5\% & 1.7\% & 82.5\% & 9.6\% \\
& \textbf{DistilRoBERTa} & \textbf{41.2\%} & \textbf{35.0\%} & \textbf{65.6\%} & \textbf{38.1\%} \\
& NRC-Intensity & 36.2\% & 3.3\% & 65.9\% & 19.8\% \\
& NRC-Label & 35.0\% & 26.7\% & 59.5\% & 30.8\% \\
\hline
\multirow{4}{*}{AAPL}
& Baseline & 10.7\% & 0.0\% & 97.0\% & 5.4\% \\
& \textbf{DistilRoBERTa} & \textbf{27.7\%} & \textbf{40.0\%} & \textbf{66.2\%} & \textbf{33.8\%} \\
& NRC-Intensity & 7.7\% & 35.0\% & 67.5\% & 21.3\% \\
& NRC-Label & 15.0\% & 4.0\% & 67.1\% & 9.5\% \\
\hline
\multirow{4}{*}{AMZN}
& Baseline & 50.8\% & 0.0\% & 66.1\% & 25.4\% \\
& \textbf{DistilRoBERTa} & \textbf{50.0\%} & \textbf{37.0\%} & \textbf{41.0\%} & \textbf{43.5\%} \\
& NRC-Intensity & 36.7\% & 38.6\% & 41.4\% & 37.6\% \\
& NRC-Label & 36.7\% & 20.0\% & 40.2\% & 28.3\% \\
\hline
\multirow{4}{*}{Average}
& Baseline & 26.3\% & 0.6\% & 81.9\% & 13.5\% \\
& \textbf{DistilRoBERTa} & \textbf{39.6\%} & \textbf{37.3\%} & \textbf{57.6\%} & \textbf{38.5\%} \\
& NRC-Intensity & 26.9\% & 25.6\% & 58.3\% & 26.2\% \\
& NRC-Label & 28.9\% & 16.9\% & 55.6\% & 22.9\% \\
\hline
\end{tabular}
\label{tab:Llama_emotion_results}
\end{table}

\begin{figure}
    \centering
    \includegraphics[width=1\linewidth]{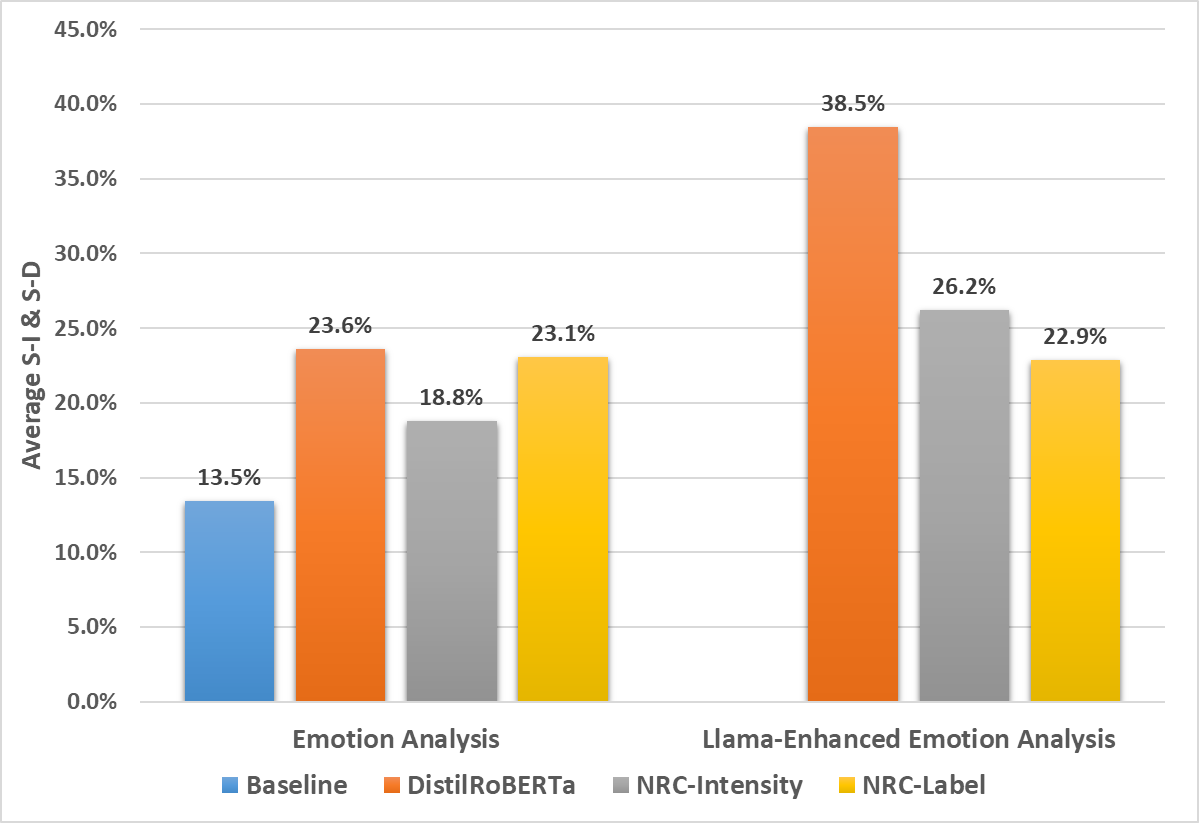}
    \caption{Overall Average Accuracy for Predicting Significant Increase and Decrease Movements Across 3 Stocks}
    \label{fig:average_SI_SD_comparison}
\end{figure}

\noindent
Table~\ref{tab:Llama_emotion_results} presents the results of the stock prediction method using Llama-enhanced emotion analysis methods, including DistilRoBERTa, NRC-Intensity, and NRC-Label. DistilRoBERTa achieves the best performance, with the overall Average S-I and S-D accuracy rising from 23.6\% (with Emotion Analysis) to 38.5\% (with Llama-enhanced Emotion Analysis), while maintaining a stable accuracy of 57.6\%. NRC-Intensity improves from 18.8\% to 26.2\%, with stable accuracy of 58.3\%. NRC-Label slightly declines from 23.1\% to 22.9\%, with stable accuracy of 55.6\%. Overall, DistilRoBERTa demonstrates the most balanced and robust performance. Figure~\ref{fig:average_SI_SD_comparison} illustrates the overall average accuracy for predicting significant increase and decrease movements across TSLA, AAPL, and AMZN stocks. Among all tested models, the DistilRoBERTa-based model consistently outperforms others. Without Llama-enhanced emotion analysis, it achieves the highest accuracy at 23.6\%, surpassing the baseline model and both NRC-based approaches. With LLaMA-enhanced emotion analysis, the DistilRoBERTa-based model enables the LSTM model to achieve 38.5\% accuracy, significantly outperforming the baseline across all three stocks. This improvement is statistically significant, with $p-value < 0.05$.

\section{\uppercase{Conclusions}}
\label{sec:conclusion}
\noindent
This paper proposes a novel framework for significant stock price movement prediction by integrating emotion features extracted from social media with historical price data. We applied the Meta’s Llama 3.1-8B-Instruct model to preprocess tweets before calculating emotion features using three emotion analysis methods: a DistilRoBERTa-based Emotion Classification model, NRC Emotion Intensity Lexicon, and NRC Emotion Label Lexicon. These features were combined with stock prices and used to train an LSTM model that takes the previous day's data as input to predict the following day's significant stock price movement. Experimental results show that incorporating emotion features improves predictive performance compared to using only historical stock prices; the DistilRoBERTa-based Emotion Classification Model configuration increased average significant increase and decrease accuracy from 13.5\% (baseline) to 23.6\% with emotion analysis, and further to 38.5\% with Llama-enhanced emotion analysis. These findings highlight the effectiveness of emotion analysis applied to social media data in financial forecasting and the added value of leveraging large language models to enhance emotion analysis.

 While our study demonstrates the utility of emotion-based features for predicting significant stock price movements, recent findings~\cite{wsj2024sentiment} indicate that consumer sentiment does not always align directly with market trends. To better capture short-term market direction, future research could combine emotion analysis with technical indicators and financial news signals to improve the accuracy of short-term stock price movement predictions.

\section*{\uppercase{Code Availability}}

\noindent The source code for this paper is available at: 
\href{https://github.com/anv0101/stock-prediction}{https://github.com/anv0101/stock-prediction}

\section*{\uppercase{Acknowledgements}}

\noindent We utilized OpenAI’s ChatGPT ~\cite{openai2024gpt4o}, a large language model, to revise the author's written text with the aim of improving its clarity, grammar, and style.


{\small
\bibliographystyle{apalike}
\bibliography{references}
}

\end{document}